\pdfoutput=1
\documentclass[11pt]{article}
\usepackage[preprint]{acl}

\usepackage[nott]{newtxtext}
\usepackage{newtxmath}

\usepackage{amsmath}
\allowdisplaybreaks[1]
\usepackage{subcaption} %
\usepackage{graphicx,xcolor}  %
\usepackage{url}

\defcitealias{juilland_etal_1971_frequency}{Juilland}
\newcommand{\citefstposs}[1]{\citetalias{#1}'s \citeyearpar{#1}}

\usepackage{booktabs}

\usepackage{relsize} %
\newcommand{\vect}[1]{\mathbf{#1}}

\newcommand{\RobustSingle}{%
    \raisebox{1.75pt}{\colorbox{darkgray}{\color{white}%
        \raisebox{-1.75pt}%
        [2pt][0pt]{\makebox[2pt]{%
            \rotatebox{15}{\textsmaller[2]{$\bigstar$}}%
        }}%
    }}%
}
\newcommand{\RobustWithLogF}{%
    \rotatebox{15}{\textsmaller[2]{$\bigstar$}}%
}

\title{Dispersion Measures as Predictors of Lexical Decision Time, Word Familiarity, and Lexical Complexity}

\author{%
    Adam Nohejl \quad Taro Watanabe\\[1ex]
    Nara Institute of Science and Technology\\[1ex]
    $\{\texttt{nohejl.adam.mt3},\;\texttt{taro}\}\texttt{@is.naist.jp}$
}

\begin{document}

\maketitle
\begin{abstract}
Various measures of dispersion have been proposed to paint a fuller picture of a word's distribution in a corpus, but only little has been done to validate them externally. We evaluate a wide range of dispersion measures as predictors of lexical decision time, word familiarity, and lexical complexity in five diverse languages. We find that the logarithm of range is not only a better predictor than log-frequency across all tasks and languages, but that it is also the most powerful additional variable to log-frequency, consistently outperforming the more complex dispersion measures. We discuss the effects of corpus part granularity and logarithmic transformation, shedding light on contradictory results of previous studies.

\end{abstract}

\section{Introduction}
\label{sec:intro}

Measures of dispersion have been proposed in corpus linguistics to complement frequency, a measure of central tendency. While a word's frequency tells us how common the word is in the whole corpus, its dispersion tells us how evenly it is spread. For instance, the words \emph{very} and \emph{yeah}, or \emph{came} and \emph{data} may have similar overall frequencies, but \emph{yeah} and \emph{data} would likely have lower dispersions, as they are specific to a certain register or domain.

The conceptually simplest dispersion measure is the range: the number of corpus parts in which a word occurs. The parts may be of different granularity and function, e.g.\ individual texts, authors, domains, or registers. In the TUBELEX corpus \cite{nohejl_etal_2024_film}, which is based on YouTube videos, our example words have the following frequencies (in thousands of occurrences) and ranges (in thousands of YouTube channels):
    \emph{very}: 332 and 35;
    \emph{yeah}: 333 and 19;
    \emph{came}: 64 and 17;
    \emph{data}: 64 and 7, confirming our expectations.

The number of texts in which a word appears was used to organize pedagogical word lists as early as in \citeyear{keniston_1920_common} by \citeauthor{keniston_1920_common} and mentioned as ``range'' by \citet{thorndike_1921_teacher}. %
\citet{gries_2008_dispersions} lists thirteen more advanced dispersion measures that have been proposed over the decades, often theoretically motivated or considering intuitive interpretability.
What is critically missing, as \citet{gries_2008_dispersions} also argues, is external validation.

We aim to bridge this gap between theoretical corpus dispersion research, psycholinguistics, and NLP applications, with a comprehensive evaluation of dispersion measures on five languages, three tasks, and three levels of corpus part granularity. Two of the tasks predict psycholinguistic data, lexical decision time (LDT) and word familiarity, and one is an NLP task, lexical complexity prediction.

\section{Related Research}

\citet{adelman_etal_2006_contextual} evaluated log-range
on English word naming and LDT, concluding that log-range\footnote{\citet{adelman_etal_2006_contextual} call range ``contextual diversity'', avoiding the terms ``dispersion'' and ``range'', which is arguably confusing \citep{hollis_2020_delineating,gries_2020_analyzing}, but does not detract anything from the practical value of their study.} is a better predictor than log-frequency.

\citet{brysbaert_new_2009_moving} replicated the results of \citeauthor{adelman_etal_2006_contextual} on the SUBTLEX-US subtitle corpus. Most later studies on film subtitles reached similar conclusions \citep{
    cai_brysbaert_2010_subtlex,%
    dimitropoulou_etal_2010_subtitle,%
    keuleers_etal_2010_subtlex,%
    pham_etal_2019_constructing,%
    boada_etal_2020_subtlex%
}, but a few of them did not find a statistical significant difference on individual datasets \citep{
    vanheuven_etal_2014_subtlex,%
    mandera_etal_2015_subtlex,%
    vanheuven_etal_2023_subtlex%
}. All of these studies investigated only log-range across subtitle files (typically thousands of files corresponding to films or show episodes) as a dispersion measure, and evaluated it on LDT or word naming times, essentially replicating \citeauthor{brysbaert_new_2009_moving}'s \citeyearpar{brysbaert_new_2009_moving} study on other languages. %

\citet{gries_2010_dispersions} evaluated multiple dispersion measures on English word naming and LDT data. The study did not reach a conclusive result on the two datasets. %

\citet{gries_2021_most} experimented with log-frequency, word length, and multiple dispersion measures as features for a random forest model of English auditory LDT. %

\section{Examined Measures}
\label{sec:dm}

As we have noted, both frequency and range are commonly log-transformed to achieve better correlation with psycholinguistic variables. Since our evaluation will include words not present in the corpus, which would result in undefined values ($\log 0$), we take the following steps to examine the log-transformation of all examined measures and frequency: Using $n =$ number of corpus parts, we apply smoothing in the form $(dn + 1) / (n + 1)$ to each dispersion measure $d$ if we log-transform it. Therefore, with a slight abuse of notation, we always use ``$\log d$'' in the following text to refer to $\log ((dn + 1) / (n + 1))$. For (log-)frequency, we always use Laplace-smoothed frequency \cite{brysbaert_diependaele_2013_dealing}.

We use all measures in forms appropriate for unequally sized corpus parts and normalized to the range $[0, 1]$, adapting the Gini index and Gries's DP to $d = 1 - d^\ast$ from their original formulas $d^\ast$, so that high values indicate high dispersion. We use the following variables, given a word $w$:
$n$ is the number of corpus parts;
$v_i$ is the number of occurrences of $w$ in part $i$;
$k_i$ is the number of tokens in part $i$;
$s_i = k_i / \smallsum\vect{k}$ is the proportion of part $i$;
$r_i = v_i / \smallsum\vect{v}$ is the proportion of occurrences of $w$ in part $i$;
$p_i = v_i / k_i$ is the relative frequency of $w$ in part $i$, i.e.\ frequency normalized per part;
$q_i = p_i / \smallsum\vect{p}$ is the frequency normalized per part and per word. For each variable $x_i$ indexed by corpus part, we understand $\vect{x} = (x_i)_{i=1}^n$ as the corresponding vector with sum $\smallsum\vect{x} = \sum_{i=1}^n x_i$, mean $\mu_\vect{x} = \smallsum\vect{x} / n$, and standard deviation $\sigma_\vect{x} = \sqrt{\sum_{i=1}^n (x_i - \mu_\vect{x})^2/n}$.

We examine the following dispersion measures:
\begin{gather}
\textrm{Range } R= \frac{\sum_{i=1}^n [ v_i > 0 ]}{n}\label{eq:range}\\
\textrm{Gini index }G = 1 - \frac{\sum_{i=1}^n \sum_{j=1}^n \left|q_i-q_j\right|}{2n}\\
\textrm{\citefstposs{juilland_etal_1971_frequency} }D = 1 - \frac{\sigma_\vect{p}}{\mu_\vect{p}\sqrt{n-1}}\label{eq:d}\\
\textrm{\citeposs{lyne_1985_vocabulary} }D_3 = 1 - \frac{\sum_{i=1}^n\left(r_i-s_i\right)^2}{4}\label{eq:d3}\\
\textrm{\citeposs{gries_2008_dispersions} DP} = 1 - \frac{\sum_{i=1}^n\left|r_i-s_i\right|}{2}\label{eq:dp}\\
\textrm{\citeposs{rosengren_1971_quantitative} } S = \frac{\left(\sum_{i=1}^n\sqrt{q_i}\right)^2}{n}\label{eq:s}\\
\textrm{\citeposs{carroll_1970_alternative} }D_2 = \frac{-\sum_{i=1}^n q_i \log q_i}{\log n}
\end{gather}

Regardless of the formulas above, we define each measure as 0 for words missing from the corpus.\footnote{
    This is in line with the formulas for range, $S$, and $D_2$, which would give 0 for a zero frequency word. $D$, $D_3$, and $\textrm{DP}$ would be undefined, and the Gini index would give 1.
}

Gini index is the discrete variant of the well-know index of inequality. It was proposed as a dispersion measure independently by \citet{murayama_etal_2018_word}, as $\textrm{Word GINI} = - \log G$, and by \citet{burch_etal_2017_measuring}, as $D_A$. The latter was adjusted to unequally sized parts, with a slightly different normalization from our $G$, by \citet{egbert_etal_2020_lexical}. We investigate $G$ and $\log G$ using the formula given by \citet{glasser_1962_variance}, which reduces computation time to $O(n\log n)$.\footnote{
    We also use sparse arrays to represent $\vect{q}$, resulting in $O(m\log m)$ time, where $m$ is the number of non-zero elements of $\vect{q}$. The sparseness of frequency vectors $\vect{q}$ grows with the number of corpus parts, keeping computation time reasonable.
}

As far as we can tell, ``distributional consistency'' proposed by \citet{zhang-etal-2004-distributional}, is simply equal to Rosengren's $S$ \eqref{eq:s}.\footnote{
    This seems to have escaped the attention of \citet{zhang-etal-2004-distributional} and Gries. Gries only noted that it gives the same numerical result in an example scenario \citep{gries_2008_dispersions} and appears similar in cluster analysis \citep{gries_2010_dispersions}.
}

Finally, we observe that inverse document frequency (idf) and variation coefficient (vc) can be expressed as linear functions of log-range \eqref{eq:range} and Juilland's D \eqref{eq:d}, respectively, and therefore do not need to be examined separately in terms of linear correlation:
\begin{align}
\textrm{idf} &= \log \frac{1}{R}s = - \log R\\
\textrm{vc} &= \frac{\sigma_\vect{p}}{\mu_\vect{p}} = \sqrt{n-1}\,(1-D)
\end{align}

\section{Evaluation}

We evaluate the measures on TUBELEX \citep{nohejl_etal_2024_film}, a large YouTube subtitle corpus for English, Chinese, Spanish, Indonesian, and Japanese. Word frequency in TUBELEX was already demonstrated to achieve correlation with psycholinguistic variables on par with or superior to film subtitle corpora \citep{nohejl_etal_2024_film}. TUBELEX also provides three levels of linguistically valid corpus parts: videos, channels, and categories (tens of thousands, thousands, and 15 parts respectively). We use TUBELEX in its default tokenization.

For evaluation, we use the same datasets for LDT (3 languages), word familiarity (5 languages), and lexical complexity (3 languages) as were used for extrinsic evaluation of TUBELEX log-frequency by \citet{nohejl_etal_2024_film}. Word familiarity and lexical complexity differ from the commonly employed LDT or word naming tasks by being based on subjective ratings as opposed to reaction time, while the lexical complexity used in this case differs from the other data by being rated by non-native speakers or a mix of natives and non-natives. %

We evaluate the dispersions in two scenarios: as single predictors, and as one of two predictors, the other one being log-frequency. In both cases, we measure adjusted $R^2$ \citep{ezekiel_1930_methods}:
\begin{equation}
    R_\textrm{a}^2 = 1-(1-R^2)\,\frac{n-1}{n-p-1}\label{eq:adjr2}
\end{equation}
where $n$ is the number of examples (dataset size) and $p$ is the number of variables (1 or 2). We compute $R^2$ (coefficient of determination) for linear least squares (multiple) regression fitted to the whole dataset, which allows us to interpret it as measure of (multiple) correlation strength.\footnote{
    Using $R^2$ instead of Pearson's (multiple) correlation coefficient $r$ ($R$) allows us to ignore the different polarity of the tasks (rare words have low familiarity but high complexity). The adjustment is appropriate for comparing different numbers of independent variables. 
}
\begin{table}[t]
\small
\setlength{\tabcolsep}{4pt}
\begin{subtable}{\linewidth}
\centering
\begin{tabular}{lccc}
\toprule
Dispersion & \multicolumn{3}{c}{$\overline{\Delta R_\textrm{a}^2}$ of $\log d$ vs. $d$ (\#Datasets: $\Delta R_\textrm{a}^2 > 0$)} \\
\cline{2-4}
Measure $d$  & Videos & Channels & Categories \\
\midrule
Range & $\phantom{-}\mathbf{0.356}$ (11) & $\phantom{-}\mathbf{0.329}$ (11) & $-0.060$ (1)\phantom{0} \\
Gini Index & $\phantom{-}\mathbf{0.361}$ (11) & $\phantom{-}\mathbf{0.340}$ (11) & $-0.000$ (5)\phantom{0} \\
Juilland's $D$ & $-0.230$\phantom{ (00)} & $-0.213$\phantom{ (00)} & $-0.202$\phantom{ (00)} \\
Gries's DP & $-0.095$\phantom{ (00)} & $-0.102$\phantom{ (00)} & $-0.169$\phantom{ (00)} \\
Rosengren's $S$ & $\phantom{-}\mathbf{0.362}$ (11) & $\phantom{-}\mathbf{0.341}$ (11) & $-0.273$\phantom{ (00)} \\
Carroll's $D_2$ & $-0.218$\phantom{ (00)} & $-0.176$ (1)\phantom{0} & $-0.241$\phantom{ (00)} \\
Lyne's $D_3$ & $-0.112$\phantom{ (00)} & $-0.089$ (1)\phantom{0} & $-0.043$\phantom{ (00)} \\
\midrule
\multicolumn{4}{c}{Frequency (for comparison): $\mathbf{0.389}$ (11)} \\
\bottomrule
\end{tabular}
\caption{Dispersion measures as single predictors.\label{subtab:logwo}}
\end{subtable}

\vspace{\baselineskip}

\begin{subtable}{\linewidth}
\centering
\begin{tabular}{lccc}
\toprule
Dispersion & \multicolumn{3}{c}{$\overline{\Delta R_\textrm{a}^2}$ of $\log d$ vs. $d$ (\#Datasets: $\Delta R_\textrm{a}^2 > 0$)} \\
\cline{2-4}
Measure $d$ & Videos & Channels & Categories \\
\midrule
Range & $\phantom{-}\mathbf{0.016}$ (8)\phantom{0} & $\phantom{-}\mathbf{0.023}$ (10) & $-0.010$ (3)\phantom{0} \\
Gini Index & $\phantom{-}\mathbf{0.003}$ (4)\phantom{0} & $\phantom{-}\mathbf{0.002}$ (5)\phantom{0} & $\phantom{-}\mathbf{0.002}$ (6)\phantom{0} \\
Juilland's $D$ & $-0.006$ (3)\phantom{0} & $-0.006$ (3)\phantom{0} & $-0.010$ (1)\phantom{0} \\
Gries's DP & $-0.002$ (3)\phantom{0} & $\phantom{-}0.000$ (7)\phantom{0} & $-0.005$ (3)\phantom{0} \\
Rosengren's $S$ & $\phantom{-}\mathbf{0.009}$ (5)\phantom{0} & $\phantom{-}\mathbf{0.009}$ (6)\phantom{0} & $-0.018$ (1)\phantom{0} \\
Carroll's $D_2$ & $-0.010$ (5)\phantom{0} & $-0.006$ (4)\phantom{0} & $-0.014$ (1)\phantom{0} \\
Lyne's $D_3$ & $-0.002$ (3)\phantom{0} & $-0.004$ (2)\phantom{0} & $-0.001$ (4)\phantom{0} \\
\bottomrule
\end{tabular}
\caption{\centering Two predictors: dispersion measure and log-frequency.\label{subtab:logw}}
\end{subtable}
\caption{Mean improvement in $R_\textrm{a}^2$ of the log-transformed measure over the non-log-transformed (number of datasets of total 11 with positive improvement, if any, in parentheses). Cases where logarithm improves $R_\textrm{a}^2$  by at least 0.001 are in bold.\label{tab:log}}
\end{table}

We predict mean LDT from three studies: the English Lexicon Project
\citep{balota_etal_2007_english}, restricted to lower-case words following the approach of Brysbaert and New \citep{brysbaert_new_2009_moving}; the MELD-SCH database \citep{tsang_etal_2018_meld} of simplified Chinese words; and SPALEX \citep{aguasvivas_etal_2018_spalex} for Spanish.
For English and Chinese, we use the published mean LDT. SPALEX only provides raw participant data, which we process by removing times out of the range [200~ms, 2000~ms] \citep{aguasvivas_etal_2018_spalex}, and computing the means. We predict mean word familiarity from five databases:
Chinese familiarity ratings \citep{su_etal_2023_familiarity},
English MRC lexical database \citep{coltheart_1981_mrc,coltheart_wilson_1987_mrc}, 
Indonesian lexical norms \citep{sianipar_etal_2016_affective},
Japanese word familiarity ratings for reception \citep{asahara-2019-word},
and Spanish lexical norms \citep{guasch_etal_2016_spanish}.
Lastly, we predict lexical complexity for English, Spanish and Japanese using the evaluation sets of the MultiLS dataset \citep{shardlow-etal-2024-bea}. In total, we are evaluating on 11 datasets (task-language combinations).

\subsection{To Log or Not to Log}

\begin{figure*}[t]
\centering
\vspace{-1pt}%
\includegraphics[width=\textwidth]{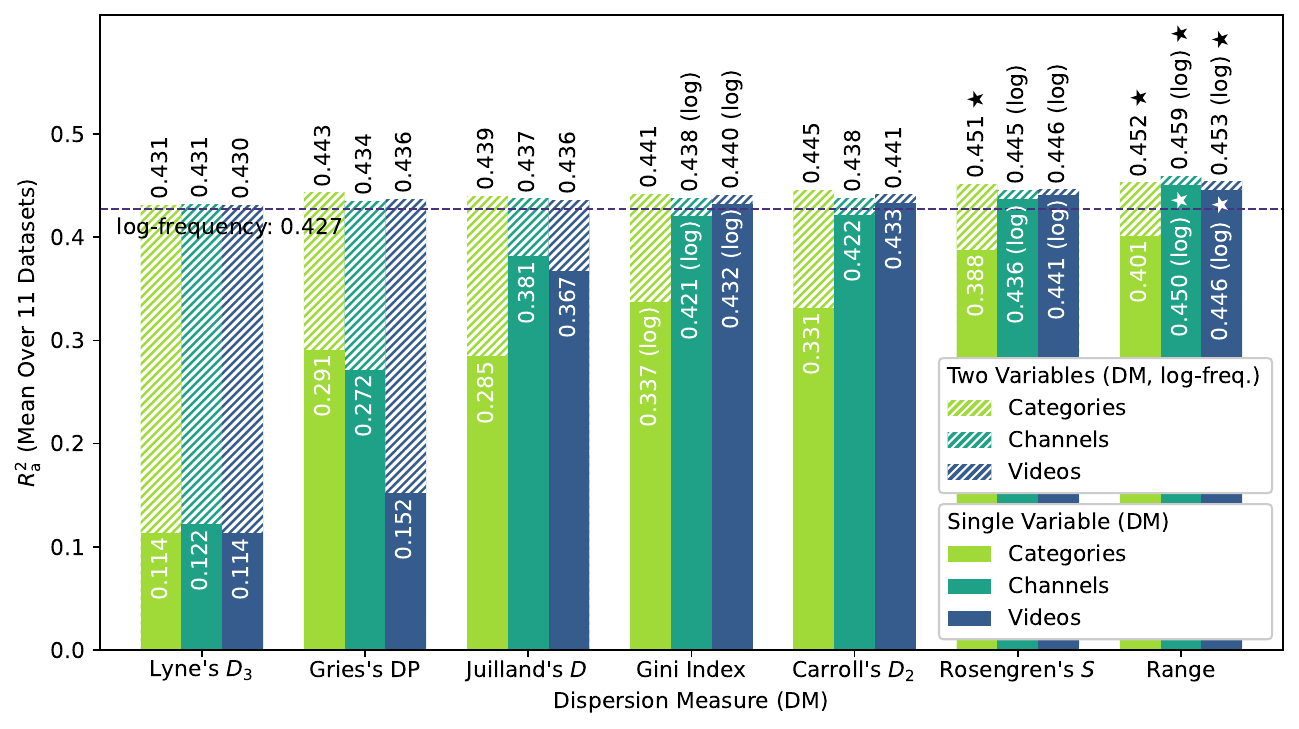}

\vspace{-0.5\baselineskip} %
\caption{Mean $R_\textrm{a}^2$ computed over 11 datasets for each dispersion measure, part granularity, and prediction with/without log-frequency as a second variable, where ``(log)'' indicates log-transformed measures. Stars indicate robust predictors, namely: \RobustSingle{} single predictors that were not significantly ($p < 0.001$) worse than log-frequency for any dataset, and \RobustWithLogF{} predictors that, when used with log-frequency, improved the prediction by $\Delta R_\textrm{a}^2 \geq 0.01$ for at least 8 of 11 datasets.}
\label{fig:disp_adj_r2}
\end{figure*}

In \autoref{tab:log}, we compare each dispersion measure with its log-transformed version. Perhaps surprisingly, which one is a better predictor does not depend solely on the measure, but also on corpus part granularity and whether the measure is used as a single predictor or with log-frequency. 

When used as single predictors (\autoref{subtab:logwo}), the logarithmic transformation benefits range, Gini index, and Rosengren's $S$, resulting in stronger correlations on all 11 datasets -- but only if videos and channels are used as parts. For all three measures, the difference between using and not using log-transformation is critical (0.340 to 0.362), comparable to that between log-frequency and frequency (0.389).

When dispersion measures are employed along with log-frequency (\autoref{subtab:logw}), applying logarithm is moderately beneficial for the same measures as above and for the Gini index for categories, but the improvements are not robust across datasets.

We will report and discuss each measure with logarithm applied or not applied according to these results.

\subsection{Results}

As shown in \autoref{fig:disp_adj_r2} (solid bars), the only dispersion measure robustly stronger than log-frequency as a predictor of LDT, word familiarity, and lexical complexity is log-range for channels and videos. %
Although it does not come near in correlation strength, Gries's DP is worth noting as the only measure performing the best with categories (the coarsest part granularity).

When used along with log-frequency, the following measures result in particularly robust improvements (in decreasing order): log-range for channels, log-range for videos, range for categories, and Rosengren's $S$ for categories, as shown in \autoref{fig:disp_adj_r2} (hatched bars).

\section{Discussion}

We extended the previous results of \citet{adelman_etal_2006_contextual} and film subtitle studies (e.g.\ \citealp{brysbaert_new_2009_moving}), which showed that log-range predicts LDT better than log-frequency, to word familiarity and lexical complexity prediction. More importantly, we found that the viability of range as a single predictor depends on (1) a fine corpus part granularity, i.e.\ channels or videos in the case of TUBELEX, and (2) the log-transformation. This explains the low correlation achieved using only non-log-transformed range in some studies, e.g.\ \citet{baayen_2010_demythologizing}. %
When dispersion is used along with log-frequency, log-range for videos and channels (fine parts) are still the best choices, followed by range and Rosenberg's $S$, both non-log-transformed and based on categories (coarse parts).

These finding offer a guideline for choosing dispersion measures as model variables, based on the corpus parts are available. The previous studies that we know of have not compared multiple part granularities of a single corpus.

Besides three levels of granularity, our evaluation encompassed 11 datasets 
(task-language combinations). As the results generally agreed across datasets, we did not report them individually. For instance, the robust single predictors (marked \RobustSingle{} in \autoref{fig:disp_adj_r2}) were significantly better than log-frequency on most datasets and not significantly different on two to three of them. We focused on the general, not the insignificant exceptions. This highlights the importance of evaluation on multiple datasets and puts into perspective the insignificant differences between log-range and log-frequencies on individual datasets reported in a few previous studies \citep{
    vanheuven_etal_2014_subtlex,%
    mandera_etal_2015_subtlex,%
    vanheuven_etal_2023_subtlex%
}. %

We believe that linear regression, which we used for analysis, gives more widely applicable and interpretable results than rank correlation \citep{gries_2010_dispersions}\footnote{
    Rank correlation is an appropriate evaluation method for applications that require only ranking, but it obscures the different ``shapes'' of dispersion metrics. %
}
or random forests \citep{gries_2021_most}\footnote{
    With enough training data a random forest may be more fitting for a practical application, but caution is needed when using it as an evaluation tool. The experiment in \citet{gries_2021_most} used the full data for both training and testing. Moreover, the features optimal for a random forest and large training data may not perform well in other scenarios.
}.
We hope that future investigations of what we have called ``exceptions'' or less ``widely applicable'' bring deeper insights into specific use cases and interactions with different data and granularities.

Our results are immediately applicable to NLP tasks that have relied on frequencies for modeling words perceived as common or simple, such as language learning applications or lexical simplification. Range data for TUBELEX, which we have used, is readily available as \texttt{channels} and \texttt{videos} in its word lists, and for most SUBTLEX language mutations as ``contextual diversity'' (CD), all based on corpus parts of comparable granularity.

\section*{Acknowledgments}

We are grateful to an anonymous reviewer of the paper by \citet{nohejl_etal_2024_film} for encouraging us to explore other dispersion metrics, given the curious result achieved by the Gini index.

\bibliography{dispersion-bib}

\end{document}